\documentclass[letterpaper, 10 pt, conference]{ieeeconf}
\usepackage[utf8]{inputenc}
\usepackage{cite}

\usepackage{amsmath}
\usepackage{amssymb}
\usepackage{hyperref}
\usepackage{relsize}%
\usepackage{graphicx}
\usepackage{subfig}
\usepackage{xcolor}
\usepackage[utf8]{inputenc}
\usepackage[english]{babel}
\usepackage{xurl}

\usepackage{algorithmicx}
\usepackage[ruled,vlined, linesnumbered]{algorithm2e}
\usepackage{comment} 
\newcommand{\ignore}[1]{}
\usepackage{color} 

\usepackage{multirow}
\usepackage[T1]{fontenc}

\makeatletter
\newcommand{\removelatexerror}{\let\@latex@error\@gobble}

\IEEEoverridecommandlockouts

\title{\LARGE \bf Autonomous Highway Merging in Mixed Traffic Using Reinforcement Learning and Motion Predictive Safety Controller}

\author{Qianqian Liu$^{1}$, Fengying Dang$^{2}$, Xiaofan Wang$^{1}$, Xiaoqiang Ren$^{1}$
\thanks{$^1$Qianqian Liu is with the School of Computer Engineering and Science, Xiaoqiang Ren and Xiaofan Wang are with the School of Mechatronic Engineering and Automation, Shanghai University, Shanghai, China, {\tt\small({e-mails: qianqianliu, xqren, xfwang}@shu.edu.cn).}}
\thanks{$^2$Fengying Dang is with the Department of Mechanical Engineering, Michigan State University, East Lansing, MI 48824, USA, {\tt\small(e-mail: dangfen1@msu.edu).}}%
}

\begin{document}

\maketitle
\thispagestyle{empty}
\pagestyle{empty}

\begin{abstract}
Deep reinforcement learning (DRL) has a great potential for solving complex decision-making problems in autonomous driving, especially in mixed-traffic scenarios where autonomous vehicles and human-driven vehicles (HDVs) drive together. Safety is a key during both the learning and deploying reinforcement learning (RL) algorithms process. In this paper, we formulate the on-ramp merging as a Markov Decision Process (MDP) problem and solve it with an off-policy RL algorithm, i.e., Soft Actor-Critic for Discrete Action Settings (SAC-Discrete). In addition, a motion predictive safety controller including a motion predictor and an action substitution module, is proposed to ensure driving safety during both training and testing. The motion predictor estimates the trajectories of the ego vehicle and surrounding vehicles from kinematic models, and predicts potential collisions. The action substitution module replaces risky actions based on safety distance, before sending them to the low-level controller. We train, evaluate and test our approach on a gym-like highway simulator with three different levels of traffic modes. The simulation results show that even in harder traffic densities, the proposed method still significantly reduces collision rate while maintaining high efficiency, outperforming several state-of-the-art baselines in the considered on-ramp merging scenarios. 
The video demo of the evaluation process can be found at: \url{https://www.youtube.com/watch?v=7FvjbAM4oFw}

\end{abstract}

\IEEEpeerreviewmaketitle

\section{Introduction}\label{sec:1}
Autonomous driving has been a hot issue from the past few decades to the present. In the future, human-driven vehicles (HDVs) and autonomous vehicles will appear on the highway at the same time. Therefore, it is challenging for an autonomous vehicle in mixed traffic to react in time to a dynamic driving environment, which requires it to not only react to road objects, but also to observe the behaviors of HDVs. There are many typical scenarios on the highway, such as automated lane changing, overtaking, turning and so on. Although most of them have remarkable achievements, one important problem that still exists is highway merging into dense traffic. In addition, efficiently and safely merging plays a significant role in easing highly congested traffic and realizing a fully autonomous vehicle.

In most autonomous driving systems today, traditional model-based Model Predictive Control (MPC) can guarantee safety during learning \cite{koller2018learning}. MPC uses the system model to predict the future behavior of the system under input and output constraints, and then provides a guarantee for the security constraint problem. For example, in \cite{dixit2019trajectory}, the trajectory planning problem is solved by a MPC approach and tested on the highway overtaking tasks, showing promising performance. There have been great achievements by using MPC methods to achieve absolute safety \cite{aswani2013provably, kabzan2019learning}, however, the exploration and performance optimization are not addressed. They cannot effectively plan for the long term and rely on accurate system models.

There are massive papers applying data-driven approaches to autonomous driving tasks. Schmerling \textit{et al.} \cite{schmerling2018multimodal} demonstrated a data-driven approach to learn the interaction model from a dataset of human-human examples. They effectively used this model for online planning with parallelization on a traffic weaving scenario involving two agents. Such a method is promising but is not suitable for dense traffic scenarios. Additionally, a tremendous amount of rule-based algorithms for on-ramp merging problems have been proposed and their effectiveness is verified \cite{scarinci2014control}. Nevertheless, they cannot adapt to complicated and unexpected situations, and collision cannot be effectively avoided in difficult scenes \cite{nishitani2020deep}. 

\begin{figure}[]
  \centering
  \includegraphics[width=0.46\textwidth]{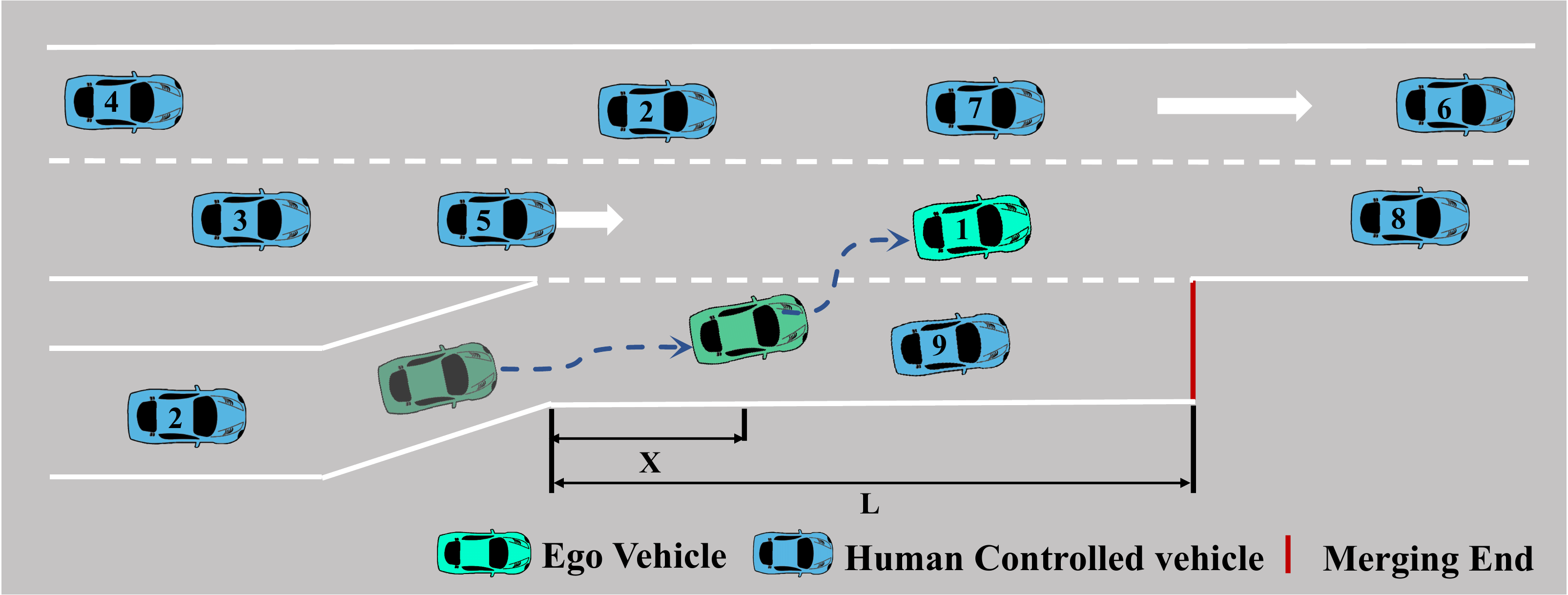}
  \caption{Example of a merging scenario in dense traffic. The main road is a multi-lane. The ego vehicle (green) and human-controlled cars (blue) randomly spawned on the lanes, either through lanes or the merging lane.}
  \label{fig:merging_scenario}
\end{figure}

Recently, reinforcement learning (RL) has received rising attention from researchers and achieves astonishing success in various other fields~\cite{mnih2013playing, silver2016mastering}. Several methods have been put forward to tackle complex autonomous driving tasks, especially in the scenario of on-ramp merging on the highway~\cite{lubars2021combining, triest2020learning, bouton2019cooperation, dong2017intention}. Lubars \textit{et al.} \cite{lubars2021combining} tackle the on-ramp merging problem through leveraging on MPC and RL and demonstrate promising performance. In \cite{triest2020learning}, a hierarchical approach is proposed to address the decision-making problem in the on-ramp merging scenario, in which a high-level decision is produced by the RL agent and actual control is executed by the low-level controller. A cooperation-aware framework is proposed in~\cite{bouton2019cooperation}, where the RL agent learns to interact with the road users and different cooperation levels are considered during training. However, most of them consider the ego vehicle to be spawned only on the through lane \cite{triest2020learning} or on the merging lane \cite{dong2017intention}. 

In this paper, we consider a more realistic on-ramp merging setting, where the ego vehicle is randomly spawned on the through or merging lanes. As shown in Fig.~\ref{fig:merging_scenario}, there are multiple through lanes and one merging lane. If the ego vehicle is driving on the upper through lane, it needs to attend to the vehicles on the current and adjacent through lanes. If the ego vehicle is on the bottom through lane, it needs to observe the merging vehicles and learn to adjust speed proactively to avoid collisions and make room for the merging vehicles. While it is more challenging when the ego vehicle is spawned on the merging lane, which requires the ego vehicle to observe vehicles both on the through lanes and merging lane. Moreover, the ego vehicle needs to safely and efficiently merge to the through lane in time to avoid traffic deadlock \cite{bouton2019cooperation}. It should also adjust its speed to adapt to the traffic on the through lanes. To simulate the real traffic conditions on the highway, we randomly spawn the ego vehicle on the lanes considering different traffic densities.

On the other hand, it is challenging for the ego vehicles to deal with the complex and dynamic driving environment, especially in the mixed-traffic cases, where estimating the motion of other road users can not only increase the driving safety but also lead to more optimal decisions of ego vehicles. In \cite{saxena2020driving}, the authors implicitly modeled the interactions between the ego vehicle and environment vehicles using a RL algorithm, i.e., Proximal Policy Optimisation (PPO). However, they cannot guarantee collision-free execution because the policy is executed without safety constraints. 

In this paper, we formulate the on-ramp merging as a Markov Decision Process (MDP) problem and solve it with an off-policy RL algorithm. Specifically, we extend the Soft Actor-Critic (SAC) \cite{haarnoja2018soft} to the discrete action settings and the agent (i.e., the ego vehicle) only makes high-level decisions including lane changes and speed shifts. With a selected high-level decision, the low-level controller then generates the corresponding steering and throttle control signals to maneuver the ego vehicle. Furthermore, a novel motion predictive safety controller is proposed to ensure the safety of the actions made by the RL agent, and to avoid over-conservation when exploring during autonomous driving. The designed safety controller consists of two components, a motion predictor and an action substitution module. More specifically, the actions generated by the RL agent are first checked by the motion predictor, in which the trajectories of the ego vehicle and environment vehicles in a considered time horizon are predicted by kinematic models. Potential collisions will be checked through overlaps in the predicted trajectories. Then the action substitution module will replace the unsafe action with the most suitable action by a designed safety room-based scheme (see section~\ref{sec:3} for details).

The main contributions of this paper are summarized as follows:
\begin{enumerate}

\item We formulate the on-ramp merging (in Fig.~\ref{fig:merging_scenario}) as a decision-making problem and solve it with an off-policy RL algorithm, i.e., discrete soft actor-critic (SAC-Discrete).

\item In addition, a motion predictive safety controller is proposed to enhance safety during autonomous driving, consisting of two main components, named the motion predictor and the action substitution module. The safety controller can significantly reduce collision rates even in complex traffic scenes during both training and evaluation.

\item Curriculum learning is adopted to enhance the learning speed for challenging tasks. Specifically, we firstly pre-train the model in the easy traffic scenes, and then transfer the learned weights to the medium and hard traffic modes. Moreover, a multi-objective reward function is further proposed to take the trade-off between safety hazards and driving efficiency.

\item Comprehensive experiments are conducted on a gym-like highway simulator under different traffic density levels, showing that the proposed approach outperforms several state-of-the-art algorithms in terms of driving safety and efficiency.
\end{enumerate}

The remainder of the paper is further organized as follows: Section~\ref{sec:2} introduces background knowledge on deep reinforcement learning (DRL) and the SAC-Discrete algorithm. Section~\ref{sec:3} describes the proposed approach for merging scenarios. Finally, the simulation environment, results, and discussions are presented in Section~\ref{sec:4}. We conclude the paper in Section~\ref{sec:5}.

\section{Preliminaries}\label{sec:2}
 
In this section, we introduce the background knowledge on DRL and the SAC-Discrete algorithm to put our proposed method in proper context. 

In a RL problem, MDP is formulated as a process by which the agent interacts with an environment, while collecting experiences and improving policy $\pi_{\phi}$ parameterized by a neural network with the parameters $\phi$. It can be defined as a tuple $(\mathcal{S}, \mathcal{A}, r, p, \gamma)$, in which $\mathcal{S}$ is the state space, $\mathcal{A}$ is the action space, $p\left(s^{\prime} \mid s, a\right)$ is the transition probability and $\gamma \in[0,1)$ is the discount factor. Furthermore, the agent receives an immediate reward through executing action $a$ in state $s$ and transits to the next state $s^{\prime}$. The main objective of the agent is to learn an optimal policy $\pi^*$ that maximizes the total discounted return $\mathcal{R}=\sum_{t=0}^{T} \gamma^{t} r_{t}$, where $r_{t}$ is the immediate reward at time step $t$.

In this paper, we extend the SAC to the discrete action space (detailed in Section~\ref{sec:3}) and output high-level decisions such as making lane changes and changing speed. Then the low-level controller generates the corresponding steering and throttle control signals to maneuver the ego vehicle. 

The three objective functions $\mathcal{J}_{Q}(\theta)$ defined in \cite{haarnoja2018soft} still hold. To modify the SAC to the discrete action space \cite{christodoulou2019soft}, the following changes are made:
\begin{enumerate}
    \item The $Q$ function is defined as $\mathcal{Q}: S \rightarrow \mathbb{R}^{|A|}$ to have the soft Q-function output the Q-value for each possible action. And the policy is $\pi: S \rightarrow[0,1]^{|A|}$ to ensure that it outputs a valid probability distribution, in which a softmax function is applied in the last layer of the network to output the discrete Q values.
    
    \item Since the action space is discrete, instead of forming a monte-Carlo estimate, the expected value can be computed directly. Thus, the state-value function is defined as:
    \begin{equation}\label{eqn:soft state-value calculation equation}
        V\left(s_{t}\right):=\pi\left(s_{t}\right)^{T}\left[Q\left(s_{t}\right)-\alpha \log \left(\pi\left(s_{t}\right)\right)\right].
    \end{equation}
    
    \item To reduce the estimated variance of the temperature loss, the temperature objective can be characterized by the following equation:
    \begin{equation}\label{eqn:temperature objective}
        J(\alpha)=\pi_{t}\left(s_{t}\right)^{T}\left[-\alpha\left(\log \left(\pi_{t}\left(s_{t}\right)\right)+\bar{H}\right)\right].
    \end{equation}
    
    \item Without the reparameterization trick, the new objective of the policy is defined as:
    \begin{equation}\label{eqn:the policy's new objective}
        J_{\pi}(\phi)=E_{s_{t} \sim D}\left[\pi_{t}\left(s_{t}\right)^{T}\left[\alpha \log \left(\pi_{\phi}\left(s_{t}\right)\right)-Q_{\theta}\left(s_{t}\right)\right]\right].
    \end{equation}
\end{enumerate}

The readers are referred to \cite{haarnoja2018soft} and \cite{christodoulou2019soft} for the derivation of the SAC-Discrete algorithm.

\section{Proposed Approach}\label{sec:3}
In this section, we formulate the on-ramp merging scene as a MDP and solve it with an off-policy RL algorithm, SAC-Discrete. Furthermore, a novel motion predictive safety controller is proposed to ensure safety during training and testing. In addition, curriculum learning is adopted to enhance the training efficiency.

\subsection{Problem Formulation}
Following is the definition of the state space, action space and reward function to formulate the MDP.

\subsubsection{\textbf{State Space}}
In addition to its state information, the ego vehicle controlled by the DRL should also know the state information of the vehicles around it.
State $\mathcal{S}$, a set of vehicle states in the on-ramp merging scenario, is defined as a $N_{\mathcal{N}_i} \times W$ matrix, where $N_{\mathcal{N}_i}$ is the number of surrounding vehicles can be observed by the ego vehicle and $W$ is the number of features representing the ego vehicle's state, containing the lateral position \textit{$x_l$}, the longitudinal position \textit{$y$}, the lateral speed \textit{$v_x$} and the longitudinal speed \textit{$v_y$} of the observable vehicle, respectively, and a binary identifier indicating if the vehicle is observable in the vicinity of the ego vehicle or not. 

Here "surrounding vehicles" is defined as the closest vehicles $\mathcal{N}_i$ relative to the ego car within a $150~m$ longitudinal distance of the current lane and adjacent lanes. As shown in Fig.~\ref{fig:merging_scenario}, if the ego vehicle is on the merging lane, we assume it can observe vehicles on the bottom through lane. We set $N_{\mathcal{N}_i} = 5$ same as the design in \cite{chen2021deep} (\textit{e.g.}, surrounding vehicles of vehicle $1$ (i.e., the ego vehicle) are vehicles $2, 5, 8$ and $9$ (HDVs)).

\subsubsection{\textbf{Action Space}}
In this paper, we follow the design in \cite{li2017explicit, chen2020autonomous} and adopt the discrete action space, which is a set with $5$ possible actions, i.e., $a_t \in \{0, 1, 2, 3, 4\}$, which represents {\it turn left, idle, turn right, speed up} and {\it slow down}, respectively.

\subsubsection{\textbf{Reward Design}}
Reward function is designed to guide the agent to learn an optimal policy considering safety and efficiency. In this paper, a multi-objective reward function is employed and defined as a linearly weighted function:
\begin{equation}\label{eqn:reward_function}
r = w_c r_c + w_s r_s + w_m r_m + w_h r_h + w_l r_l.
\end{equation}

The $w_c$, $w_s$, $w_m$, $w_h$ and $w_l$ are positive weighting parameters for collision evaluation $r_c$, stable-speed evaluation $r_s$, merging cost evaluation $r_m$, headway cost evaluation $r_h$, and lane-change cost evaluation $r_l$, respectively.
To prioritize the safety standards, we set other weights much smaller than $w_c$. The design of each evaluation term is specified as follows:
\begin{itemize}
\item The $r_c$ represents the collision evaluation, which is defined as:
    \begin{equation}\label{eqn:reward_collison}
        r_c \triangleq 
        \begin{cases}
              -1, & \text{If collision happened,} \\
              0, & \text{Otherwise}.
        \end{cases}
    \end{equation}

\item The $r_s$ is a high speed reward to evaluate the stability and efficiency during the whole task. Hence, the stable-speed evaluation is represented as:
    \begin{equation}\label{eqn:reward_speed}
        r_s \triangleq 
        \begin{cases} 
           \operatorname{clip}\left(\frac{v_{t}-{v}_{\min}}{{v}_{\max}-{v}_{\min}}, 0, 1\right), & {v}_{\min} \leq v_{{t}} \leq {v}_{\max}, \\
          0, & \text{Otherwise},
        \end{cases}
    \end{equation}

where $v_t$ is the current speed of the ego vehicle, ${v}_{\min}$ and ${v}_{\max}$ are minimum and maximum speeds, respectively.

\item To avoid deadlocks, the $r_m$ is used to penalize the waiting time on the merging lane \cite{bouton2019cooperation} and is defined as:
    \begin{equation}\label{eqn:reward_merge}
        r_m \triangleq 
        \begin{cases} 
        -\exp (\frac{-(x-L)^2}{10 L}), & \text{If stay on the merging lane}, \\
        0, & \text{Otherwise},
        \end{cases}
    \end{equation}

where $x$ is the distance navigated by the ego vehicle on the ramp and $L$ is the length of the ramp (see Fig.~\ref{fig:merging_scenario}). 

\item Headway evaluation $r_h$ time is adopted to avoid potential collisions with the preceding vehicles and is defined as:
    \begin{equation}\label{eqn:reward_headway}
        r_h \triangleq 
        \begin{cases}
        \log {\frac{d_{\text{headway}}}{t_h v_t}}, & \text{If } \log {\frac{d_{\text{headway}}}{t_h v_t}} < 0, \\
        0, & \text{Otherwise},
        \end{cases}
    \end{equation}
    
    where $d_{headway}$ is the distance headway and $t_h$ is a predefined time headway threshold. The ego vehicle will be penalized if the time headway is less than $t_h$. We set $t_h=1.2s$ in this paper according to~\cite{ayres2001preferred}.

\item To maintain driving comfort, the $r_l$ is the lane-change reward to reduce unnecessary and frequent lane changes~\cite{saxena2020driving}, which is defined as:
    \begin{equation}\label{eqn:reward_lane-change}
        r_l \triangleq 
        \begin{cases}
        -1, & \text{If change lanes}, \\
        0, & \text{Otherwise}.
        \end{cases}
    \end{equation}

\end{itemize}


\subsection{Motion Predictive Safety Controller}
To improve safety and efficiency during autonomous driving, we propose a framework for RL incorporating a motion predictive safety controller, which consists of a motion predictor and an action substitution module.


\subsubsection{Motion Predictor}
\begin{figure}[]
  \centering
  \includegraphics[width=0.48\textwidth]{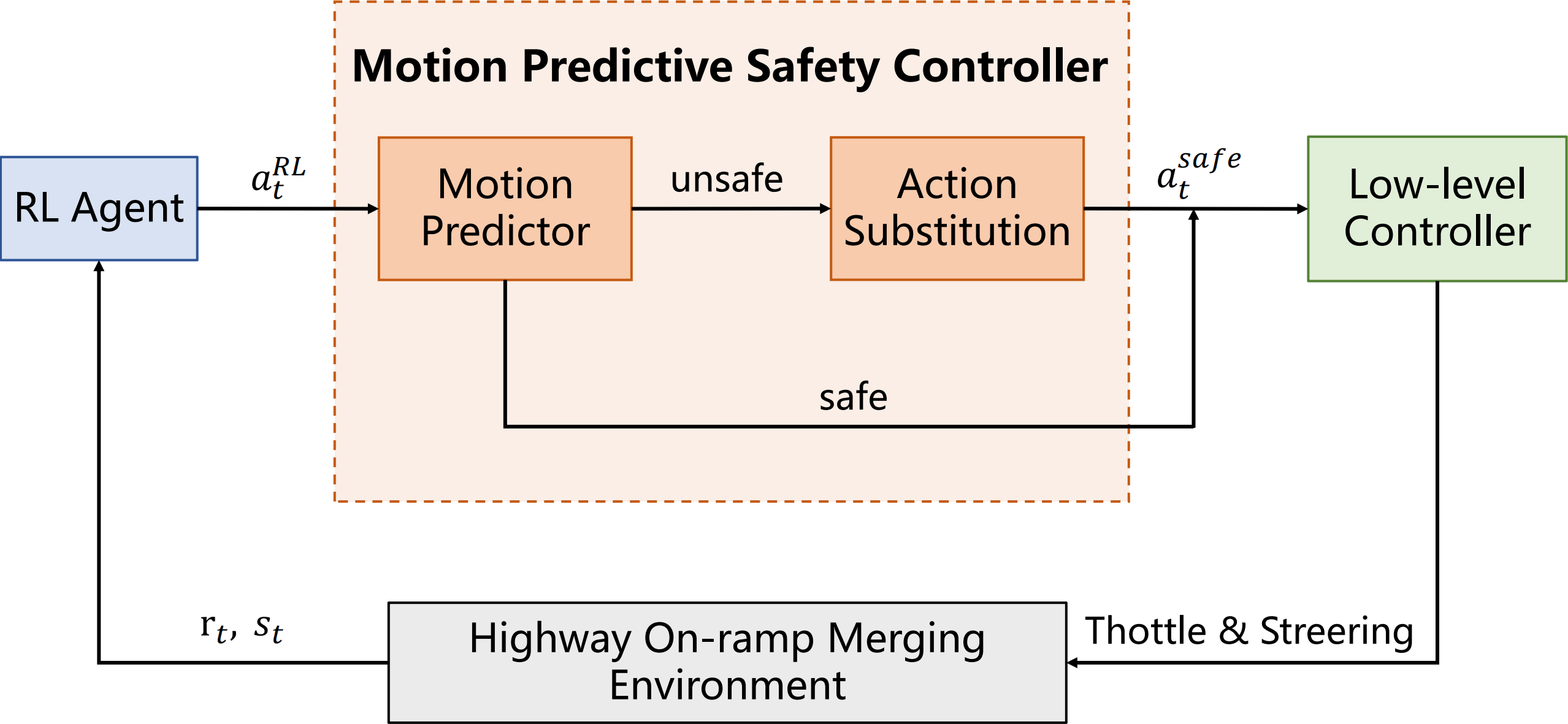}
  \caption{Structure of the motion predictive safety controller: $a$ - action, $s_t$ - state, $r_t$ - reward, ${a}_{t}^{RL}$ - action from RL, $a_t^{safe}$ - action from the motion predictive safety controller at $t$ time step, respectively.}
  \label{fig:system_structure}
\vspace{-10pt}  
\end{figure}

As shown in Fig.~\ref{fig:system_structure}, a RL agent receives the state of the current traffic scenario, and then generates an action. The motion predictor predicts the traveling trajectories in a time horizon $T_n$ of the "surrounding vehicles" $\mathcal{N}_i$ relative to the ego vehicle firstly. Then it checks whether the motion primitive caused by the exploration action will collide with its surrounding vehicles. Specifically, for HDVs, the longitudinal acceleration is predicted by the intelligent driver model (IDM) \cite{treiber2000congested}, based on the current speed and distance between vehicles in the transit system. The lateral behaviors of HDVs are predicted by the minimizing overall braking induced by lane change (MOBIL) \cite{kesting2007general}. For the ego vehicle, the high-level decisions are made by the agent with discrete action space. A collision risk can be detected if the predicted trajectory of the ego vehicle coincide with any other considered vehicles (i.e., the distance between the two sequences at any step $k$, $k=1,\cdots,T_n$ is below a specified threshold, \textit{e.g.}, the vehicle size). If no collision is detected, the exploration action will be sent directly to the low-level controller to maneuver the ego vehicle. It should be noted that IDM and MOBIL models will guarantee that there is (almost) no collision among HDVs in the on-ramp merging scenarios.

\subsubsection{Action Substitution}
If a collision with other HDVs is detected, the action made by the RL agent is determined to be unsafe, and it will be replaced by the action substitution module with a new "safe" action. With a safe action, lower-level PID controllers then produce the corresponding steering and throttle control signals for navigating the ego vehicle in the driving environment. Then the environment transits to the next state, and returns the immediate reward and next state to the agent. Specifically, the safe action is selected from the other available actions according to the following rule:

\begin{equation}\label{eqn:safe_action}
a^{\prime}_{t} = {\arg\max}_{a_t \in \mathcal{A}_{\text{available}}} \big(\min_{k \in T_n} d_{\text{sp}, k} \big),
\end{equation}
where $\mathcal{A}_{\text{available}}$ is the set of available actions at time step $t$,  $d_{\text{sp}, k}$ is the safety space at the prediction time step $k$ and is defined as:

\begin{equation}\label{eqn:safety_space}
    d_{\text{sp}, k} \triangleq
    \begin{cases}
        \min{\left|P_{\text{v}_j, t}-P_{\text{v}_e, t} \right|}, & \text{Change lanes}, \\
        P_{\text{v}_j[0], t}-P_{\text{v}_e, t}, & \text{On the through lanes}, \\
        P_{\text{v}_j[2], t}-P_{\text{v}_e, t}, & \text{On the ramp},
    \end{cases}
\end{equation}
where $P_{\text{v}_e, t}$ and $P_{\text{v}_j, t}$ are the positions of the ego vehicle and its surrounding vehicles at time step $t$, respectively, $P_{\text{v}_j[0], t}$ and $P_{\text{v}_j[2], t}$ are the positions of the preceding cars on the left and right in the surrounding vehicles at time step $t$, respectively.

\subsubsection{SAC-Discrete with Safety Controller}
\begin{figure}
\removelatexerror
\scalebox{0.95}{
\begin{algorithm*}[H]
\SetAlFnt{\small}
    \SetKwInOut{Parameter}{Parameters}
    \SetKwInOut{Output}{Outputs}
\caption{SAC-Discrete with Motion Predictive Safety Controller}
\label{algo:safe_SACD_with_safety_controller}
\SetAlgoLined
\Parameter{$Q_{\theta_{1}}, Q_{\theta_{2}}, \bar{Q}_{\theta_{1}}, \bar{Q}_{\theta_{2}}, \pi_{\phi}, T_n, T_s, T_u, M$.}
\Output{$\theta_{1}, \theta_{2}, \phi$.}
\vspace{0.2em}
\hrule
\vspace{0.2em}
Initialise $Q_{\theta_{1}}: S \rightarrow \mathbb{R}^{|A|}, Q_{\theta_{2}}: S \rightarrow \mathbb{R}^{|A|}, \pi_{\phi}: S \rightarrow[0,1]^{|A|}$\;

Initialise $\bar{Q}_{\theta_{1}}: S \rightarrow \mathbb{R}^{|A|}, \bar{Q}_{\theta_{2}}: S \rightarrow \mathbb{R}^{|A|}$\;

$\bar{\theta}_{1} \leftarrow \theta_{1}, \bar{\theta}_{2} \leftarrow \theta_{2}$\;
$\mathcal{D} \leftarrow \emptyset$.

\For{$j = 0$ to $M-1$}{
    \For{$t\%T_s==0$}{
        $a_{t} \sim \pi_{\phi}\left(a_{t} \mid s_{t}\right)$\;
        find surrounding vehicles $\mathcal{N}_{\text{v}_e}$ of the ego car $\text{v}_e$\;
        predict trajectories $\zeta_{v}, v \in \text{v}_e \cup \mathcal{N}_{\text{v}_e}$ for $T_n$ time steps.

        \If{safe}{
        execute $a_{t}$\;
        $s_{t+1} \sim p\left(s_{t+1} \mid s_{t}, a_{t}\right)$\;
        $\mathcal{D} \leftarrow \mathcal{D} \cup\left\{\left(s_{t}, a_{t}, r\left(s_{t}, a_{t}\right), s_{t+1}\right)\right\}$.
        }
        
        \Else{
        update $a_{t} \leftarrow a^{\prime}_{t}$ according to Eqn.~\eqref{eqn:safe_action} and execute $a^{\prime}_{t}$\;
        replace the trajectory $\zeta_{\text{v}_e}$ with $\zeta^{\prime}_{\text{v}_e}$\;
        $s_{t+1} \sim p\left(s_{t+1} \mid s_{t}, a^{\prime}_{t}\right)$\;
        $\mathcal{D} \leftarrow \mathcal{D} \cup\left\{\left(s_{t}, a^{\prime}_{t}, r\left(s_{t}, a^{\prime}_{t}\right), s_{t+1}\right)\right\}$.
        }
    }
    \For{$t\%T_u==0$}{
        $\theta_{i} \leftarrow \theta_{i}-\lambda_{Q} \hat{\nabla}_{\theta_{i}} J\left(\theta_{i}\right)$ for $i \in\{1,2\}$\;
        $\phi \leftarrow \phi-\lambda_{\pi} \hat{\nabla}_{\phi} J_{\pi}(\phi)$\;
        $\alpha \leftarrow \alpha-\lambda \hat{\nabla}_{\alpha} J(\alpha)$\;
        $\bar{Q}_{i} \leftarrow \tau Q_{i}+(1-\tau) \bar{Q}_{i}$ for $i \in\{1,2\}$.
    }
}
\end{algorithm*}}
\end{figure}

Based on the motion predictive safety controller discussed before, Pseudo-code of the SAC-Discrete with the proposed safety controller is shown in Algorithm~\ref{algo:safe_SACD_with_safety_controller}. Firstly, initializing parameters, target network weights, and an empty replay buffer (Lines $1$--$4$). In each time step, sampling an action from the policy (Line $7$), then finding surrounding vehicles relative to the ego vehicle and predicting trajectories in a time horizon $T_n$ of them to check if the action is safe (Line $8$--$9$). If safe, the agent will execute it without other checks. Then sample transition from the environment and store the experience in the replay buffer (Lines $11$--$13$). Otherwise, the risky action will be replaced with a "safe" action by the action substitution module according to Eqn.~\eqref{eqn:safe_action}. The safe action will be taken by the agent and the corresponding transition will be sampled and saved to the replay buffer (Lines $16$--$19$). After the completion of each gradient step, the parameters of Q-function and the weights of policy network are updated using the collected experience sampled from the off-policy experience buffer. The temperature is adjusted automatically to increase the stability of the algorithm (Lines $23$--$26$). 

\subsection{Curriculum Learning}
Curriculum learning can utilize prior knowledge before training, and extract dynamics information during training \cite{jiang2015self}. So it can gradually proceed from easy to more complex scenarios in training. 

In this paper, we firstly train the algorithm on an easy traffic mode, and then leverage the learned weights both in medium and hard training modes to improve the training efficiency. Curriculum learning is particularly beneficial for safety-critical tasks such as autonomous driving, since starting with a decent model can greatly reduce the amount of potentially risky and "blind" exploration \cite{chen2021deep}. 

\section{Result And Evaluation}\label{sec:4}
In this section, we introduce the simulation environment and the designed on-ramp merging scenarios. Then we compare the proposed approach with several state-of-the-art baseline methods under different traffic densities.

\subsection{Simulation Settings and Baselines}
In order to increase the complexity of the simulation environment as shown in Fig.~\ref{fig:merging_scenario}, we define three different traffic modes by setting the different number of initial HDVs on the lanes \cite{bouton2019cooperation}, i.e., easy, medium and difficult traffic densities. Detailed settings of the simulation scenarios are shown in Table~\ref{tab:simulation_scenario}, and the hyperparameters about SAC-Discrete are specified in Table~\ref{tab:sacd_parameters}.

We compare our approach with the following state-of-the-art algorithms:
\begin{enumerate}
    \item DRL: This policy is trained using state-of-the-art DRL methods (i.e., ACKTR \cite{wu2017scalable}, PPO \cite{schulman2017proximal}, A2C \cite{mnih2016asynchronous}, SAC-Discrete \cite{christodoulou2019soft} without safety controller), and we compare the evaluation return and average speed with our approach, where SAC-Discrete is an off-policy algorithm for the discrete action space without the proposed safety controller.
    
    \item Ours (SAC-Discrete with safety controller): This policy is trained using the SAC-Discrete with the proposed motion predictive safety controller. We also conduct comparative experiments in different values of $T_n$ (i.e., $T_n=3,7,9$) to verify the impact of prediction steps, which is referred to as Safe SACD.
\end{enumerate}

\begin{table}[!ht]
\renewcommand{\arraystretch}{1.3}
\centering
\caption{Settings of the simulation scenario}
\label{tab:simulation_scenario}
\setlength{\tabcolsep}{5.2mm}{
\begin{tabular}{|cc|c}
\hline
\multicolumn{2}{c|}{\textbf{Traffic simulator terms}}        & \textbf{Value}           \\ \hline
\multicolumn{2}{c|}{Total lane length}                      & 480m            \\ \hline
\multicolumn{2}{c|}{Merge lane length}                      & 80m             \\ \hline
\multicolumn{2}{c|}{Simulation frequency}                   & 15Hz            \\ \hline
\multicolumn{2}{c|}{Policy frequency}                       & 5Hz             \\ \hline
\multicolumn{2}{c|}{Initial speed}                          & 25m/s with random noise \\ \hline
\multicolumn{1}{c|}{\multirow{3}{*}{Traffic mode}} & Easy   & 6-8 HDVs        \\ \cline{2-3} 
\multicolumn{1}{c|}{}                              & Medium & 9-12 HDVs       \\ \cline{2-3} 
\multicolumn{1}{c|}{}                              & Hard   & 13-15 HDVs      \\ \hline
\end{tabular}
}
\end{table}

\begin{table}[!ht]
\renewcommand{\arraystretch}{1.3}
\centering
\caption{Hyperparameters of SAC-Discrete}
\label{tab:sacd_parameters}
\setlength{\tabcolsep}{5mm}{
\begin{tabular}{c|c}
\hline
\textbf{SAC-Discrete terms} & \textbf{Value} \\ \hline
Num\_steps                  & 1000000        \\ \hline
Batch size                  & 256            \\ \hline
Memory size                 & 500000         \\ \hline
$\gamma$                       & 0.99           \\ \hline
Start steps                 & 1000           \\ \hline
Update interval             & 4 steps        \\ \hline
Target update interval      & 8000 steps     \\ \hline
Max episode steps           & 1000           \\ \hline
Evaluation interval         & 50 episodes    \\ \hline
\end{tabular}
}
\vspace{-10pt}
\end{table}

\subsection{Experimental Results and Discussion}

In this subsection, we evaluate the effectiveness of the proposed SAC-Discrete with the proposed motion predictive safety controller. Our comprehensive experiments consist of two main parts. In one part, we evaluate the effectiveness of the proposed safety controller for different prediction horizons $T_n$ in terms of evaluation reward and average speed. In another part, we conduct the performance comparisons between our approach and four state-of-the-art baseline methods in three different merging traffic modes.

\begin{figure}[t]
	\centering
	\subfloat[easy traffic mode]{
		\includegraphics[width=0.22\textwidth]{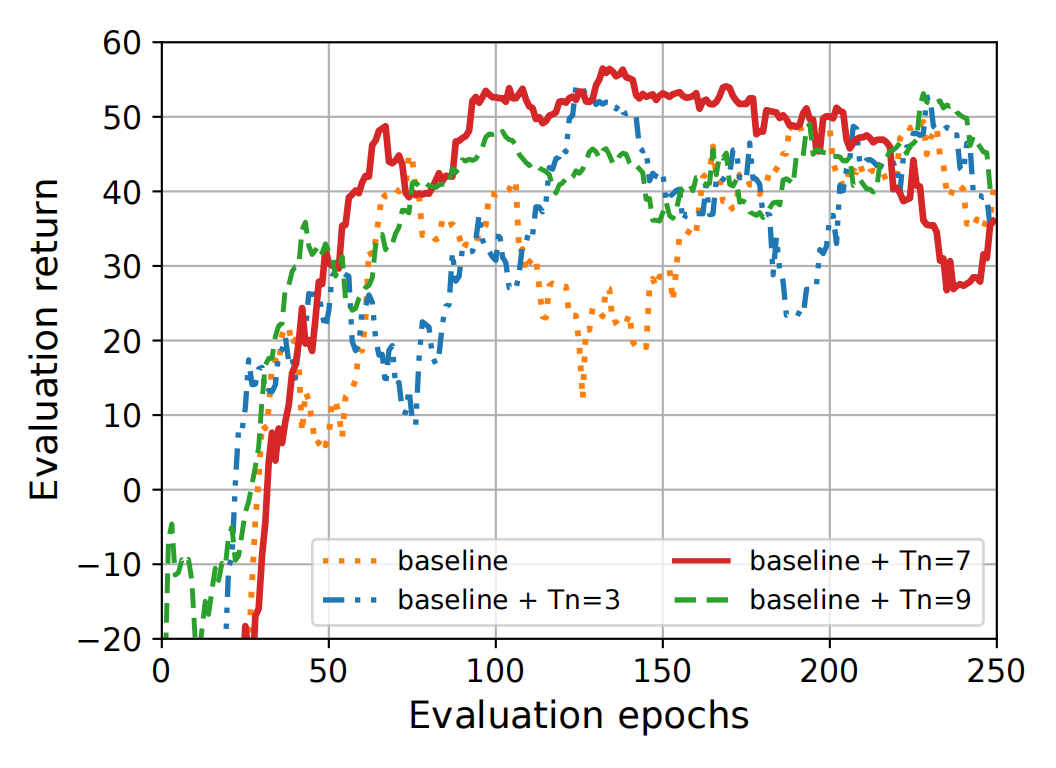}
	}
	\hfill
	\subfloat[easy traffic mode]{
		\includegraphics[width=0.22\textwidth]{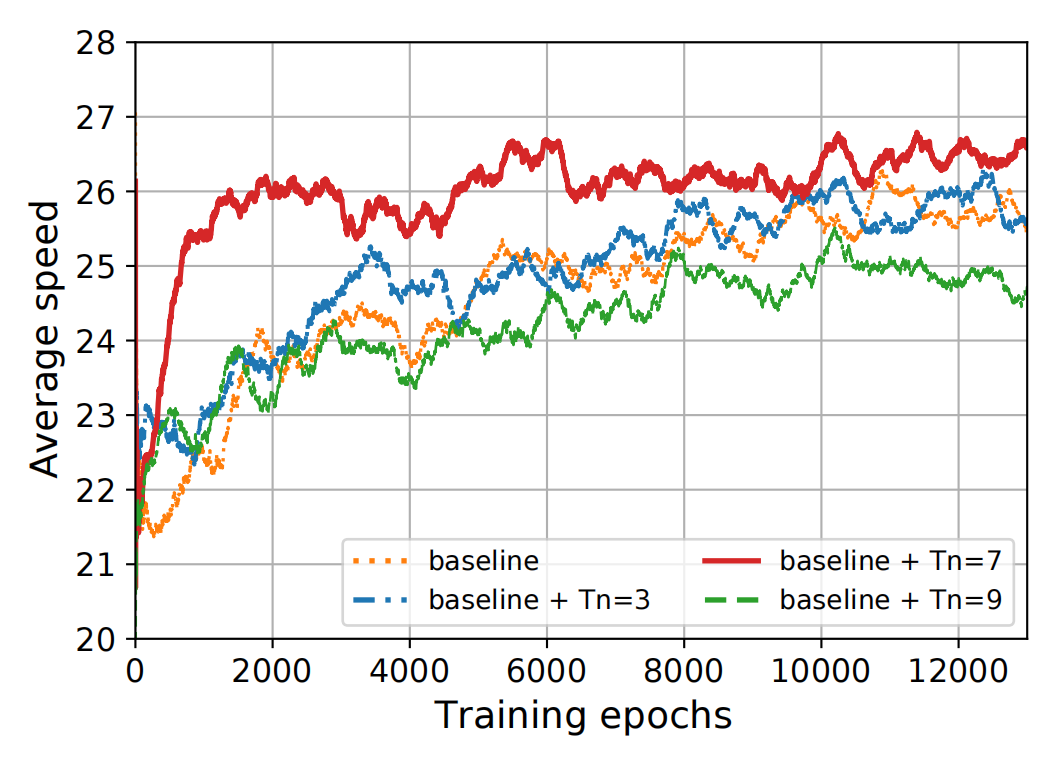}
		\label{fig:b_easy_mode_avgspeed}
	}
	\newline
	\subfloat[medium traffic mode]{
		\includegraphics[width=0.223\textwidth]{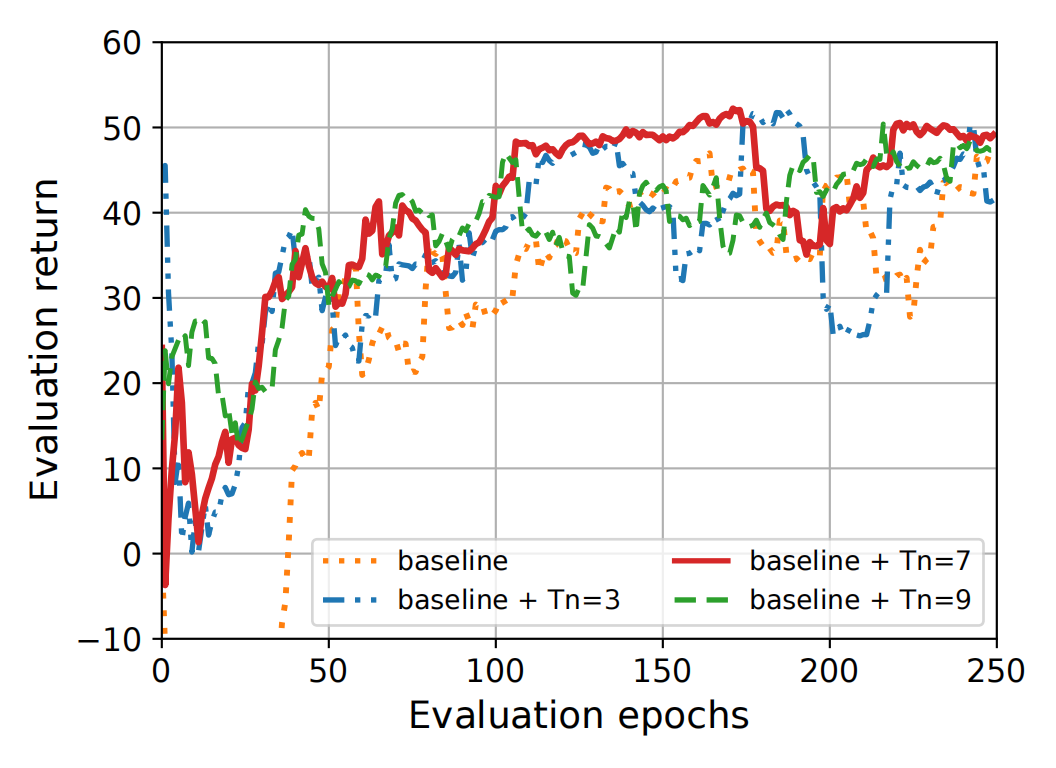}
	}
	\hfill
	\subfloat[medium traffic mode]{
		\includegraphics[width=0.22\textwidth]{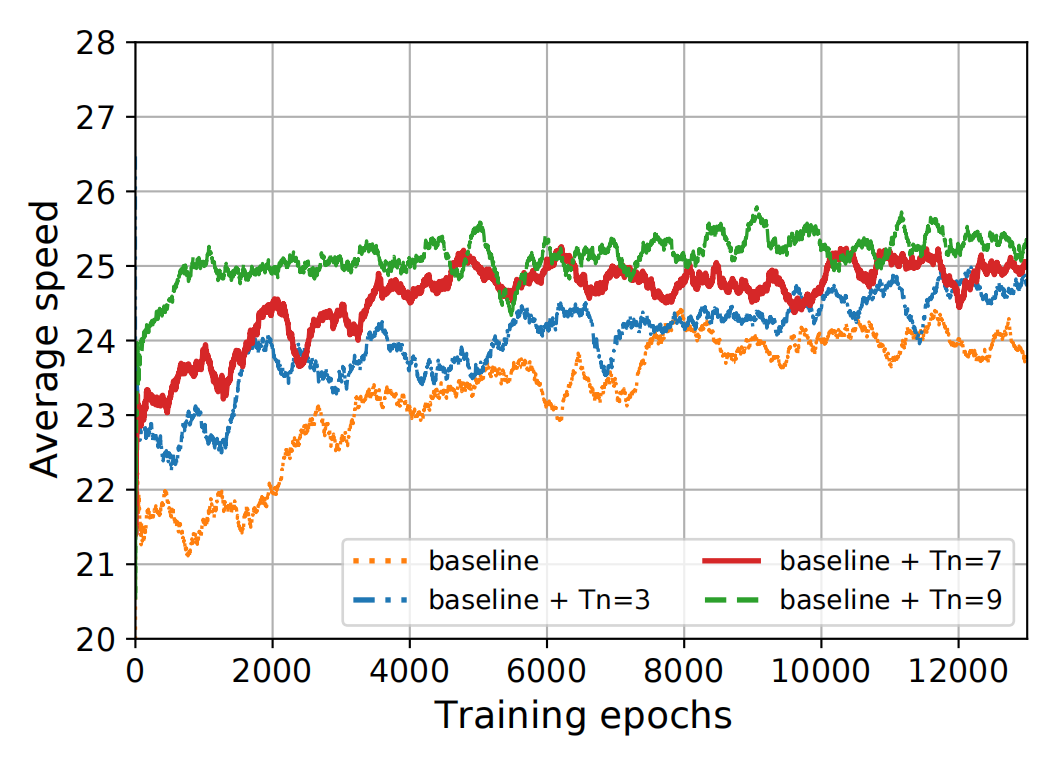}
	}
	\newline
	\subfloat[hard traffic mode]{
		\includegraphics[width=0.22\textwidth]{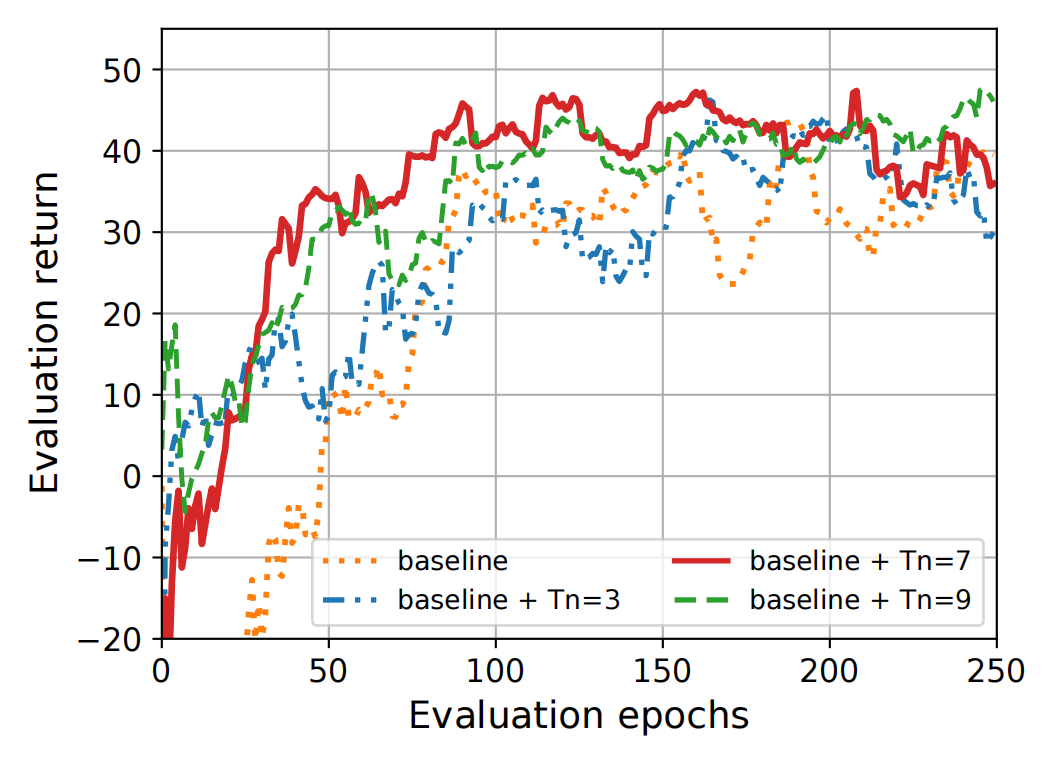}
	}
	\hfill
	\subfloat[hard traffic mode]{
		\includegraphics[width=0.22\textwidth]{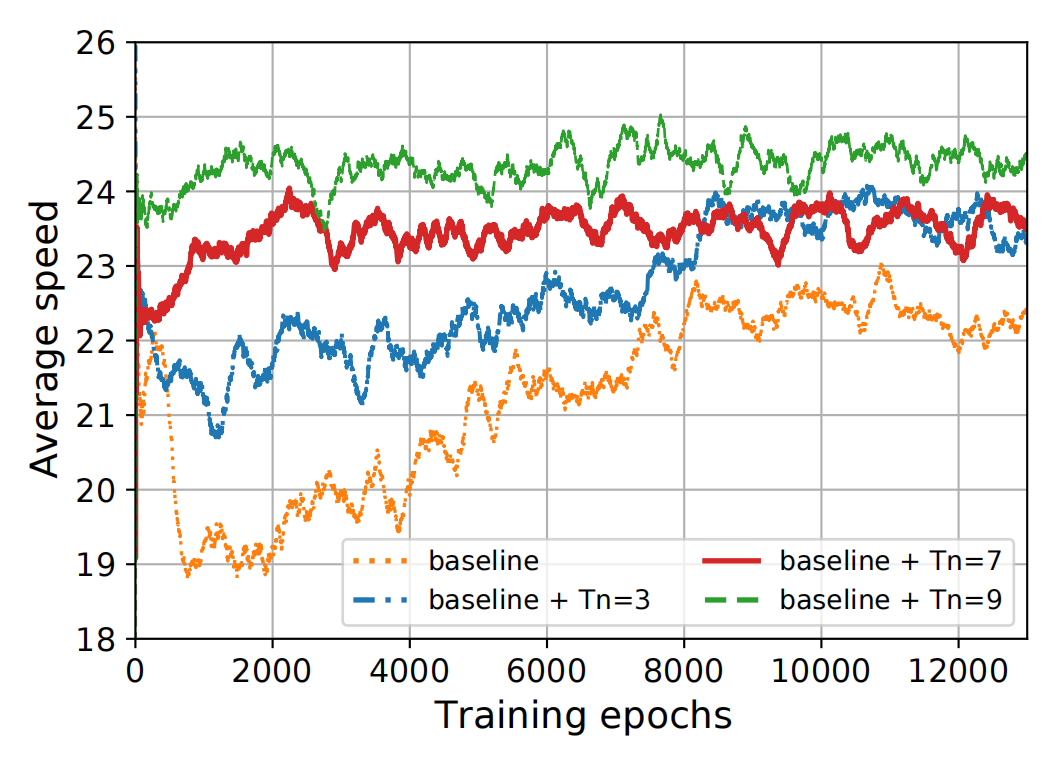}
	}
	\caption{Evaluation return and average speed comparisons for different values of $T_n$ in three traffic modes.}
	\label{fig:evaluation_reward_safety}
\end{figure}

Fig.~\ref{fig:evaluation_reward_safety} shows the evaluation return and average driving speed of the ego vehicle for the proposed safety controller under different prediction time horizons $T_n$. It is clear that the prediction horizon plays an important role in controlling the evaluation performance. The safety controller improves the evaluation return and average speed and outperforms the baseline method (i.e., the SAC-Discrete without safety controller) by a large margin. For example, the methods with the safety controller have faster convergence speed and average speed than the baseline method in the hard traffic mode, \textit{e.g.}, baseline + $T_n=9$ achieves about $24~m/s$, while the baseline method only achieves around $22~m/s$. A large prediction horizon, \textit{e.g.}, $T_n=7$ or $T_n=9$, achieves better performance in terms of learning speed and evaluation return than smaller $T_n$ since a large prediction horizon provides more predicted trajectories. Note that a large prediction horizon also brings high computational costs and delays. It is interesting to find that the average speed of the baseline + $T_n=9$ is worse than the baseline method in the easy traffic mode (shown in Fig.~\ref{fig:b_easy_mode_avgspeed}). That’s because a large prediction horizon also makes the ego vehicle behave conservatively. As shown from the test data in Table~\ref{tab:collion_safty}, we can see that the baseline method without a safety controller performs poorly in the hard traffic mode having a high collision rate of $0.17$, while baseline + $T_n=7$ can drive collision-free in all traffic modes.

\begin{table}[]
\renewcommand{\arraystretch}{1.3}
\centering
\caption{Comparison of collision rate and average speed with different prediction steps $T_n$ in the proposed approach during testing. }
\label{tab:collion_safty}
\setlength{\tabcolsep}{3.5mm}{
\begin{tabular}{c|c|c|c|c}
\hline
$T_n$                               & Metrics              & Easy & Medium & Hard \\ \hline
\multirow{2}{*}{baseline}           & collision rate       & 0.17    & 0      & 0.1    \\ \cline{2-5} 
                                    & avg. speed {[}m/s{]} & 25.35    & 24.59      & 22.45    \\ \hline
\multirow{2}{*}{$T_n=3$} & collision rate       & 0    & 0.1      & 0.07    \\ \cline{2-5} 
                                    & avg. speed {[}m/s{]} & 26.63    & \textbf{25.75}      & 23.36    \\ \hline
\multirow{2}{*}{$T_n=7$} & collision rate       & 0    & 0      & 0    \\ \cline{2-5} 
                                    & avg. speed {[}m/s{]} & \textbf{27.47}    & 25.51      & \textbf{25.01}    \\ \hline
\multirow{2}{*}{$T_n=9$} & collision rate       & 0    & 0      & 0    \\ \cline{2-5} 
                                    & avg. speed {[}m/s{]} & 26.33    & 24.99      & 24.5    \\ \hline
\end{tabular}
}
\end{table}



\begin{figure}[t]
	\centering
	\subfloat[easy traffic mode]{
		\includegraphics[width=0.22\textwidth]{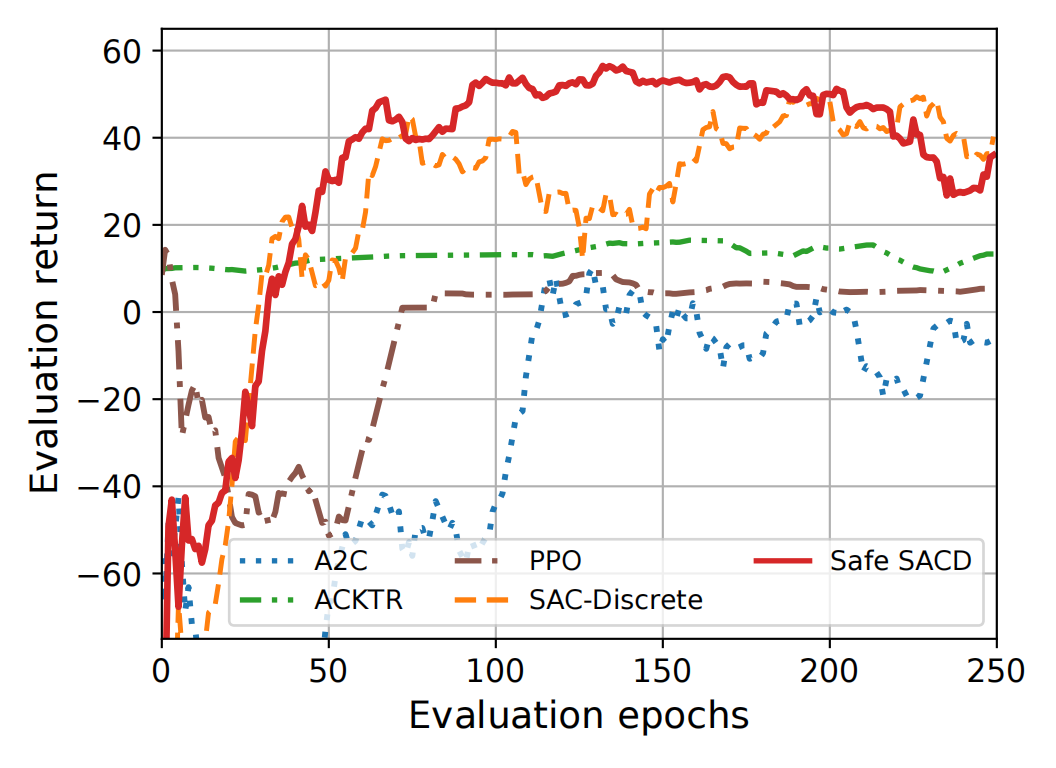}
	}
	\hfill
	\subfloat[easy traffic mode]{
		\includegraphics[width=0.22\textwidth]{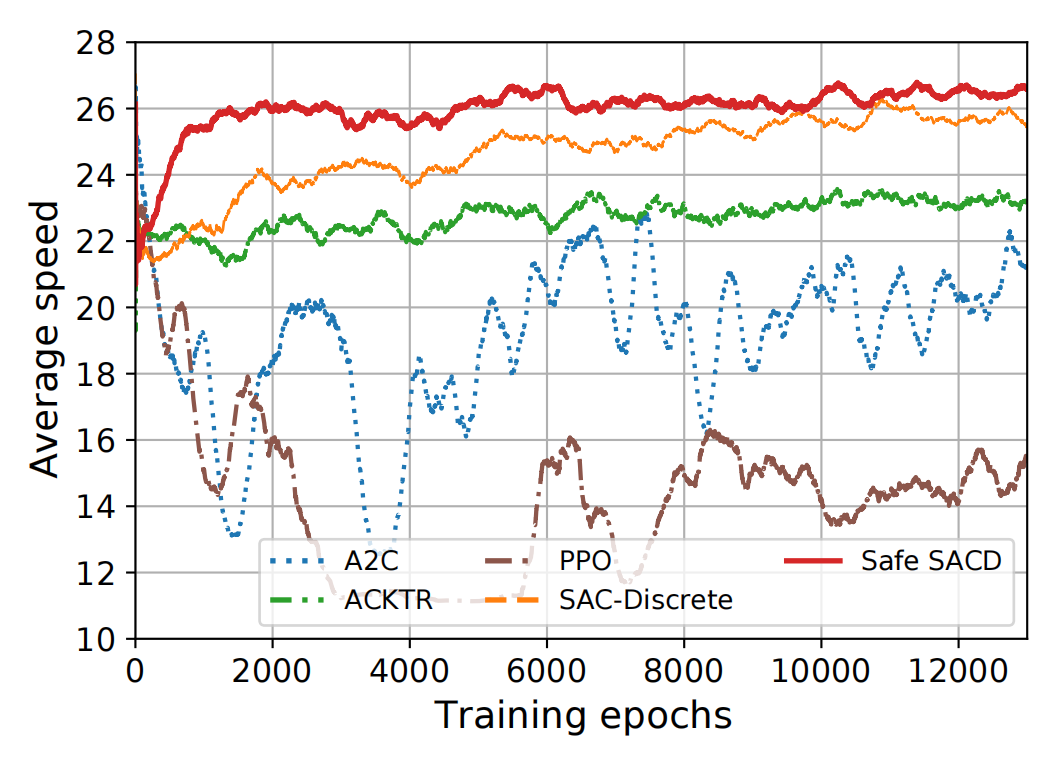}
	}
	\newline
	\subfloat[medium traffic mode]{
			\includegraphics[width=0.22\textwidth]{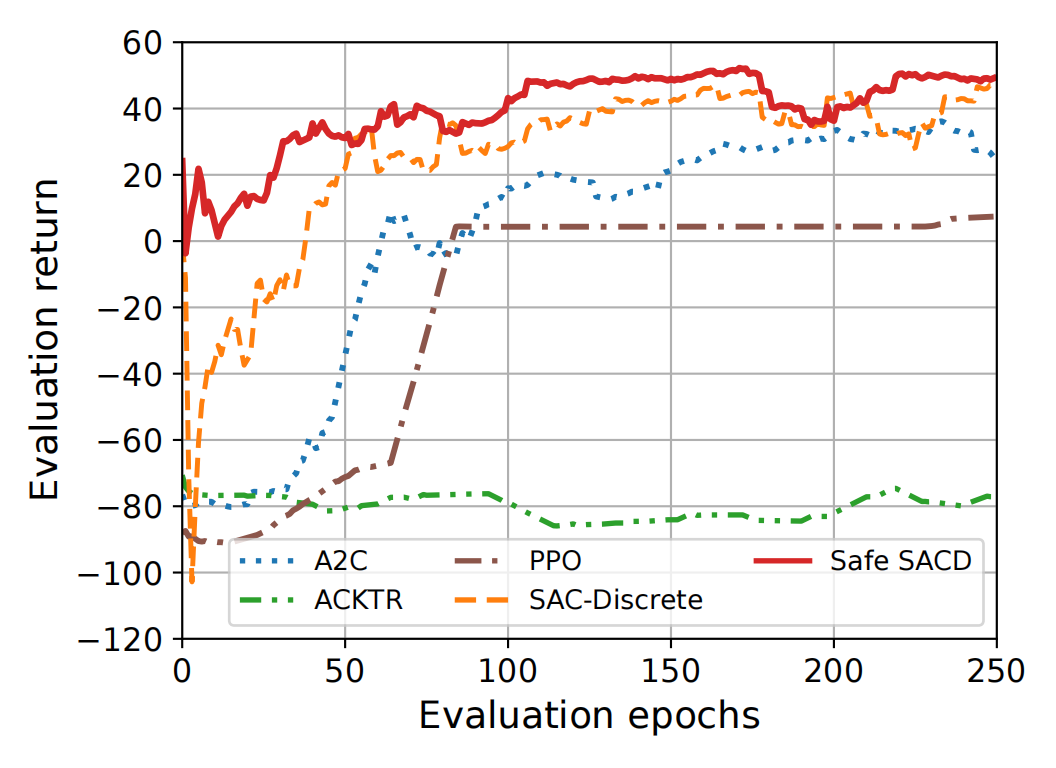}
	}
	\hfill
	\subfloat[medium traffic mode]{
			\includegraphics[width=0.22\textwidth]{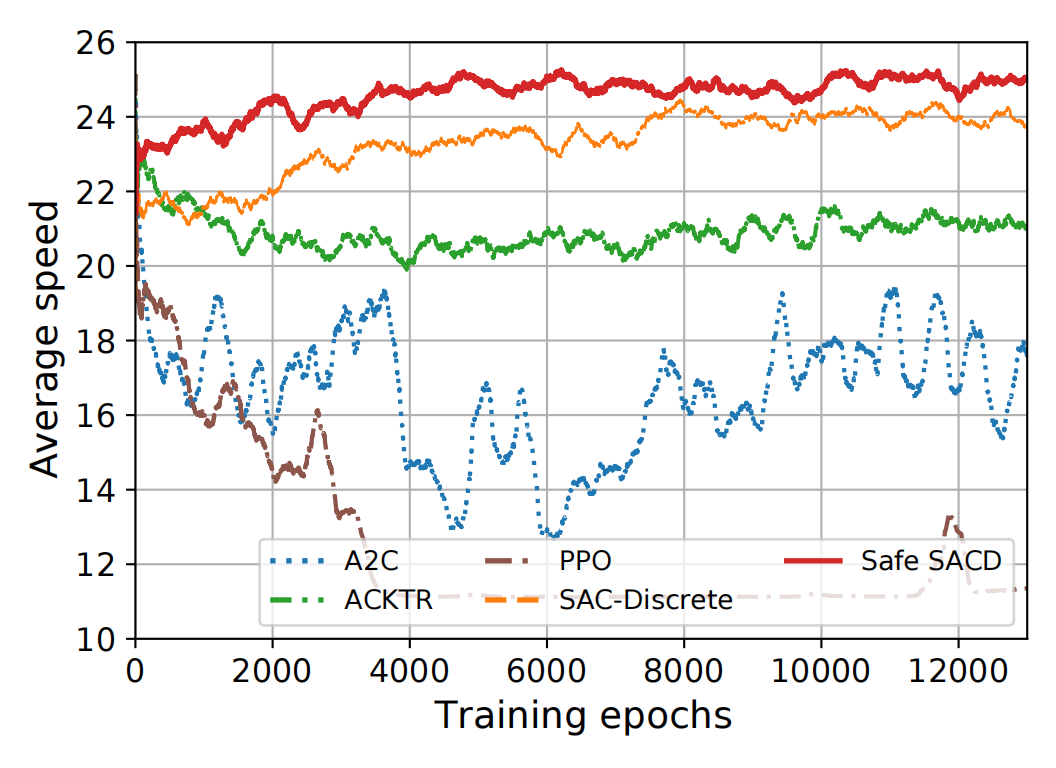}
	}
	\newline
	\subfloat[hard traffic mode]{
		\includegraphics[width=0.22\textwidth]{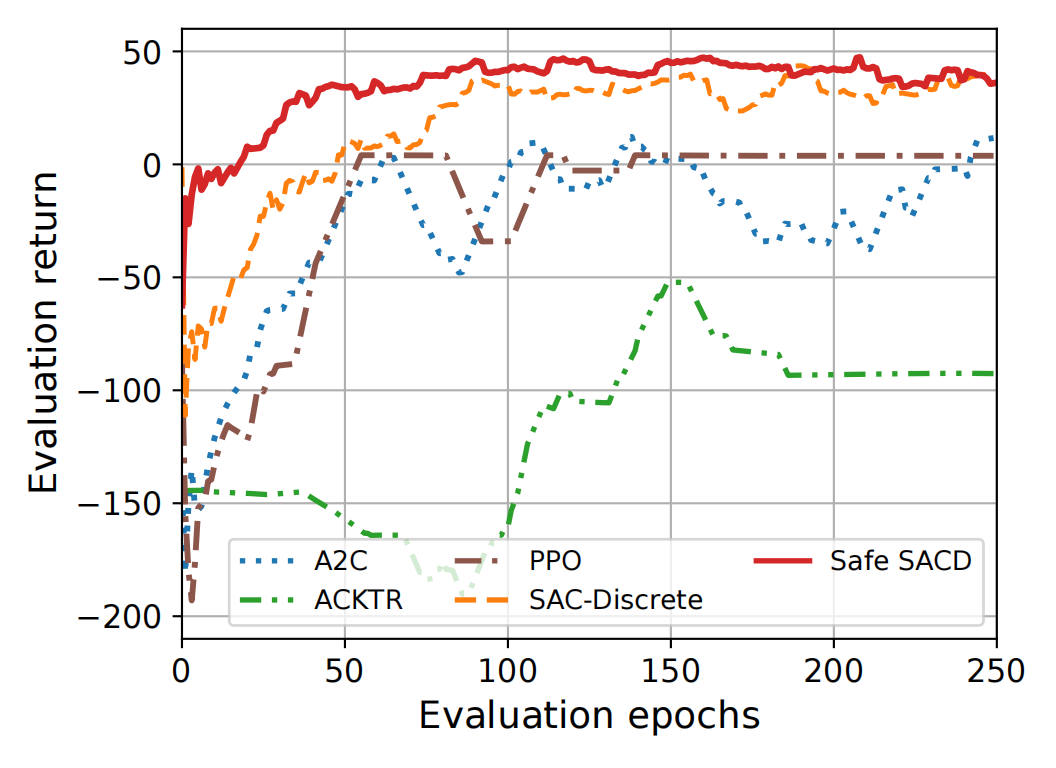}
	}
	\hfill
	\subfloat[hard traffic mode]{
		\includegraphics[width=0.22\textwidth]{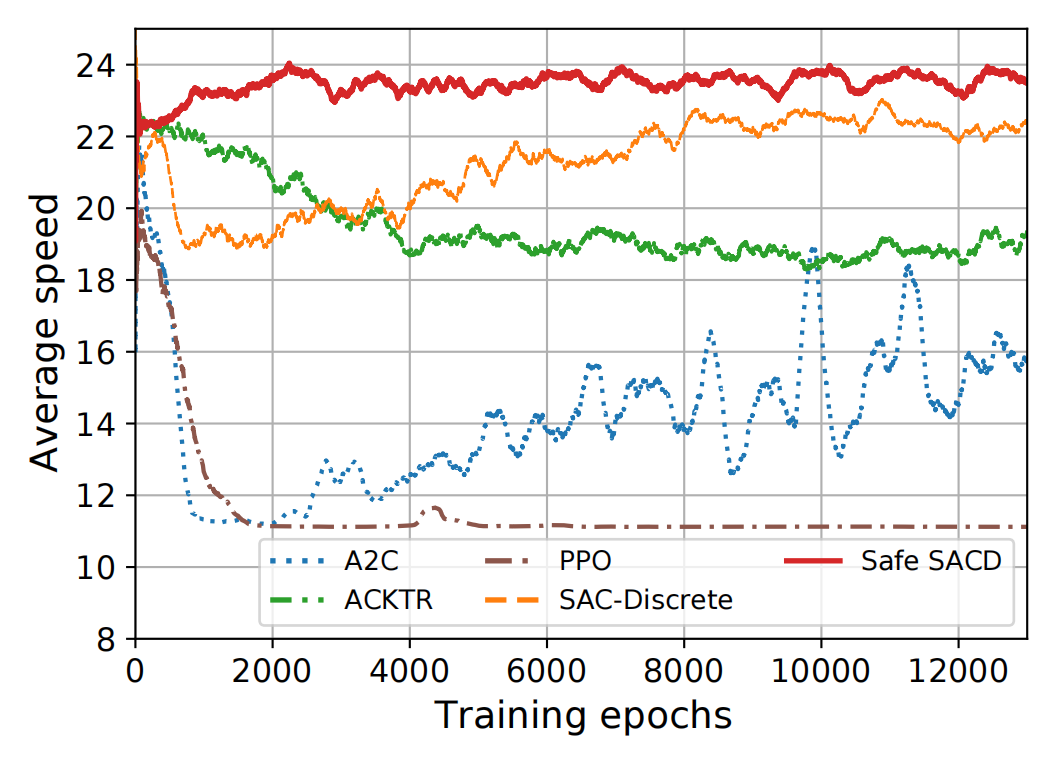}
	}
	\caption{Evaluation return and average speed comparisons between four state-of-the-art baselines (ACKTR, PPO, A2C, SAC-Discrete) and ours (Safe SACD) in three traffic modes, where ours is based on $T_n=7$.}
	\label{fig:evaluation_reward_benchmark}
\end{figure}

Fig.~\ref{fig:evaluation_reward_benchmark} shows the comparisons between the proposed approach with four state-of-the-art benchmarks. As expected, the Safe SACD shows the best and the fastest learning ability, as its training curve steadily increases and then becomes stable within a shortest time. On the other hand, the Safe SACD also archives the highest average speed than other methods. For example, it archives $24~m/s$ compared to PPO at $10~m/s$ and A2C at $16~m/s$. This is due to the proposed safety controller and curriculum learning schemes, which help improve sample efficiency and speed up the learning. After training, we test the algorithms for $30$ random episodes and Table~\ref{tab:collion_benchmark} presents the testing results of the proposed approach and benchmarks in terms of collision rate and average speed. It is clear that Safe SACD consistently achieves the best execution performance over other baseline methods in all traffic scenarios with different traffic densities. Especially, Safe SACD archives the zero-collision rate even in the most challenging case, i.e., hard traffic mode. The SAC-Discrete without the safety controller also shows good performance in the medium traffic mode while it fails in the easy and hard traffic modes with high collision rates. ACKTR also does not perform well in all the scenarios which is consistent with the training curves shown in Fig.~\ref{fig:evaluation_reward_benchmark}. It is surprising that PPO achieves relatively low collision rate, however, it has a lower average speed than Safe SACD. 

\begin{table}[]
\renewcommand{\arraystretch}{1.3}
\centering
\caption{Comparison of collision rate and average speed between the proposed approach and four state-of-the-art baselines during testing. }
\label{tab:collion_benchmark}
\setlength{\tabcolsep}{2.8mm}{
\begin{tabular}{c|c|c|c|c}
\hline
Method                                                                                             & Metrics              & Easy & Medium & Hard \\ \hline
\multirow{2}{*}{ACKTR}                                                                             & collision rate       & 0.2    & 0.4      & 0.5    \\ \cline{2-5} 
                                                                                                   & avg. speed {[}m/s{]} & 24.42    & 20.98      & 20.33    \\ \hline
\multirow{2}{*}{PPO}                                                                               & collision rate       & 0    & 0.1      & 0    \\ \cline{2-5} 
                                                                                                   & avg. speed {[}m/s{]} & 20.06    & 16.13      & 11.17    \\ \hline
\multirow{2}{*}{A2C}                                                                               & collision rate       & 0.07    & 0.2      & 0.1    \\ \cline{2-5} 
                                                                                                   & avg. speed {[}m/s{]} & 24.46    & 22.95      & 18.13    \\ \hline
\multirow{2}{*}{SAC-Discrete}                                                                      & collision rate       & 0.17    & 0      & 0.1    \\ \cline{2-5} 
                                                                                                   & avg. speed {[}m/s{]} & 25.35    & 24.59      & 22.45    \\ \hline
\multirow{2}{*}{ours} & collision rate       & 0    & 0      & 0    \\ \cline{2-5} 
                                                                                                   & avg. speed {[}m/s{]} & \textbf{27.47}    & \textbf{25.51}      & \textbf{25.01}    \\ \hline
\end{tabular}
}
\end{table}

\section{Conclusion}\label{sec:5}
In this paper, we formulate the on-ramp merging as a MDP and solve it using an off-policy RL algorithm, SAC-Discrete, equipping with a motion predictive safety controller, which includes a motion predictor and an action substitution module. The motion predictor is responsible for predicting potential collisions between the ego vehicle and surrounding vehicles. The action substitution module aims at replacing risky actions, before sending them to the low-level controller. The proposed approach consistently performs superior merging behaviors with lower collision rate and higher average speed compared to other methods at different traffic densities. Especially in the hard traffic mode, the proposed method can prosperously satisfy constraints (\textit{e.g.}, safety and efficiency) for autonomous vehicle behavior with \textit{zero} collision rate. In addition, we incorporate curriculum learning to enhance the learning efficiency of the agent in harder tasks.

Despite recent considerable achievements in DRL, its application in real-world engineering systems is still penurious. In the future work, we will focus on optimizing the safety controller to provide local guidance for the autonomous vehicle, improving it with the real human-driving data for training and a real-world driving environment for evaluation.

\bibliography{ref}
\bibliographystyle{IEEEtran}

\end{document}